%% file: main.tex
\documentclass[10pt,twocolumn,letterpaper]{article}

\usepackage[pagenumbers]{cvpr} 
\usepackage{enumitem}

\usepackage{amsfonts}    
\usepackage{amsmath}
\usepackage{booktabs}   
\usepackage{pifont}
\usepackage{amssymb}
\usepackage{amsthm}
\usepackage{xcolor}
\usepackage{multirow}
\usepackage{bm}
\usepackage{array}
\usepackage{color}
\usepackage{colortbl}
\usepackage{textcomp}
\usepackage{stfloats}
\usepackage{verbatim}
\usepackage{graphicx}
\usepackage{booktabs}
\usepackage{subcaption}
\usepackage{booktabs} 
\usepackage{arydshln} 
\usepackage{tcolorbox}
\usepackage{adjustbox}
\usepackage{CJKutf8}
\usepackage{microtype}
\usepackage{pgfplots}

\definecolor{red}{RGB}{255,44,0}
\definecolor{ired}{RGB}{229,72,72}
\definecolor{igreen}{RGB}{80,219,144}


\definecolor{arxivblue}{rgb}{0.21,0.49,0.74}
\usepackage[pagebackref,breaklinks,colorlinks,allcolors=arxivblue]{hyperref}

\def\paperID{} 

\title{Measuring the Unspoken: A Disentanglement Model and Benchmark for Psychological Analysis in the Wild}

\author{Yigui Feng\\
College of Computer Science, National University of Defense Technology\\
Deya Road 109, Changsha, 410073, Hunan, China
\and
Qinglin Wang\\
College of Computer Science, National University of Defense Technology\\
Deya Road 109, Changsha, 410073, Hunan, China
\and
Haotian Mo\\
College of Computer Science, National University of Defense Technology\\
Deya Road 109, Changsha, 410073, Hunan, China
\and
Yang Liu\\
Shien-Ming Wu School of Intelligent Engineering, South China University of Technology\\
xingye Road 777, Guangzhou, 511442, Guangdong, China
\and
Ke Liu\\
College of Computer Science, National University of Defense Technology\\
Deya Road 109, Changsha, 410073, Hunan, China
\and
Gencheng Liu\\
College of Computer Science, National University of Defense Technology\\
Deya Road 109, Changsha, 410073, Hunan, China
\and
Xinhai Chen\\
College of Computer Science, National University of Defense Technology\\
Deya Road 109, Changsha, 410073, Hunan, China
\and
Siqi Shen\\
School of Informatics Xiamen University\\
Furong Road, Xiang'an District, Xiamen, 361005, Fujian, China
\and
Songzhu Mei\\
College of Computer Science, National University of Defense Technology\\
Deya Road 109, Changsha, 410073, Hunan, China
\and
Jie Liu\\
College of Computer Science, National University of Defense Technology\\
Deya Road 109, Changsha, 410073, Hunan, China
}

\pgfplotsset{compat=1.18}

\begin{document}

\maketitle

\begin{abstract}
Generative psychological analysis of in-the-wild conversations faces two fundamental challenges: (1) existing Vision-Language Models (VLMs) fail to resolve Articulatory-Affective Ambiguity, where visual patterns of speech mimic emotional expressions; and (2) progress is stifled by a lack of verifiable evaluation metrics capable of assessing visual grounding and reasoning depth. We propose a complete ecosystem to address these twin challenges. First, we introduce Multilevel Insight Network for Disentanglement(MIND), a novel hierarchical visual encoder that introduces a Status Judgment module to algorithmically suppress ambiguous lip features based on their temporal feature variance, achieving explicit visual disentanglement. Second, we construct ConvoInsight-DB, a new large-scale dataset with expert annotations for micro-expressions and deep psychological inference. Third, Third, we designed the Mental Reasoning Insight Rating Metric (PRISM), an automated dimensional framework that uses expert-guided LLM to measure the multidimensional performance of large mental vision models. On our PRISM benchmark, MIND significantly outperforms all baselines, achieving a +86.95\% gain in micro-expression detection over prior SOTA. Ablation studies confirm that our Status Judgment disentanglement module is the most critical component for this performance leap. Our code has been opened.
\end{abstract}

\section{Introduction}
\label{sec:intro}
The aspiration to create artificial intelligence that genuinely understands human beings is a long-standing goal in computer science, holding the key to transformative applications from mental healthcare to trustworthy AI~\cite{Du2025humanlike}. The true frontier for this endeavor lies not in controlled laboratory settings, but in the complex, unconstrained dynamics of "in-the-wild" conversations—the very scenarios where human understanding is most challenged, as psychological research like Truth-Default Theory (TDT)~\cite{Levine2024relative}suggests. It is in these low-stakes interactions where the only reliable clues to a person's inner state are often the most subtle, involuntary, and purely visual "emotional leakages"~\cite{Silla2022detection}.

\begin{figure*}[t]
\centering
\includegraphics[width=0.8\textwidth]{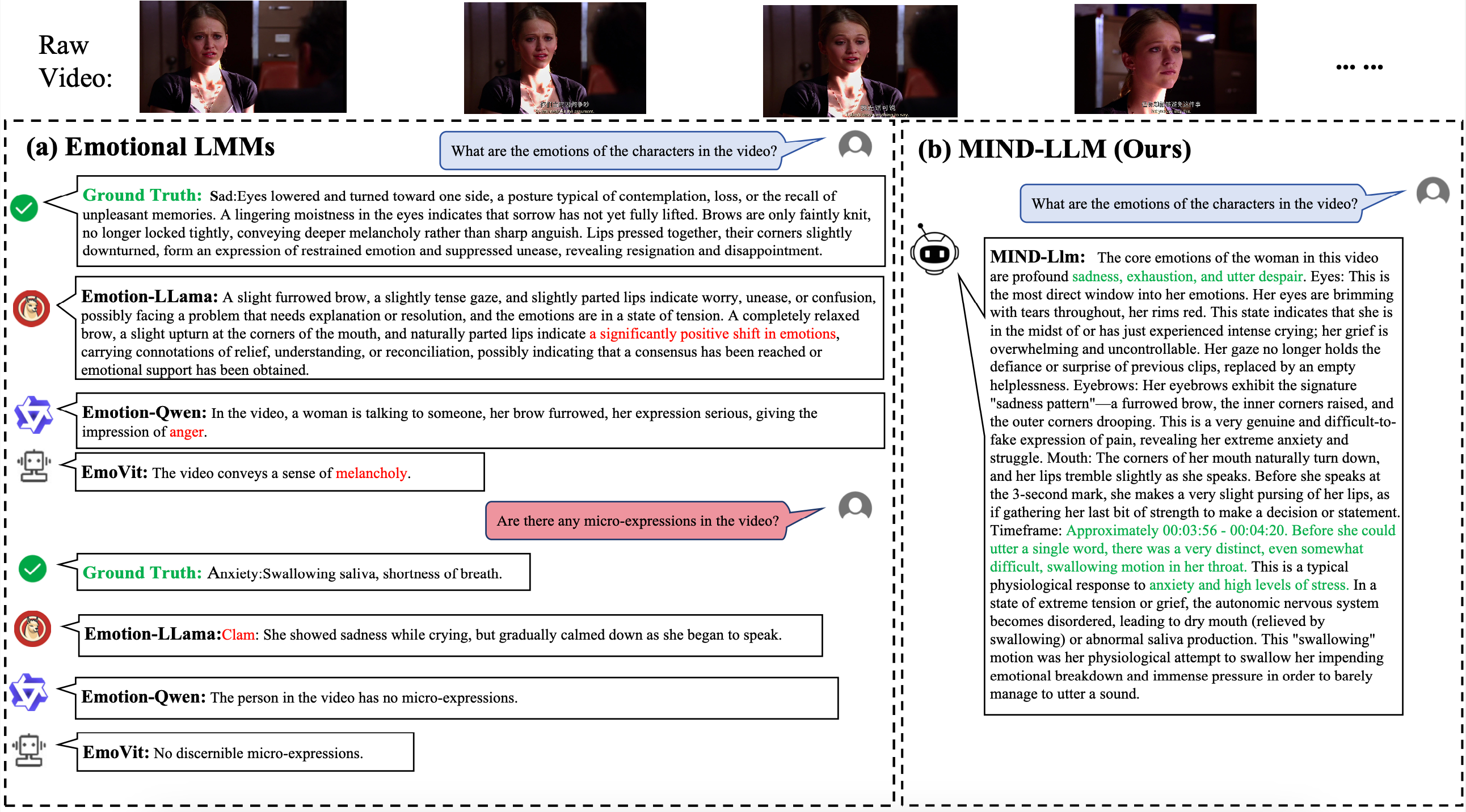} 
\caption{The motivation behind MIND (zoom in for detailed Q\&A): (a) Current state-of-the-art large multimodal models (LMMs) for emotion recognition are inaccurate for video-based emotion recognition, unable to detect and analyze micro-expressions, and have limited ability to infer a person's psychological activities. (b) In contrast, MIND effectively addresses these limitations, enabling efficient psychological analysis and profiling of individuals. Incorrect outputs are marked in red; correct outputs are marked in green.}
\label{fig1}
\end{figure*}

Capturing these visual signals is paramount, yet this presents a dual challenge. First, there is a critical technical obstacle in the visual domain which we term Articulatory-Affective Ambiguity. The same facial muscles used for emotional expression are also used for speech, creating profound ambiguity in a video-only analysis~\cite{Seiger2022perception}. Current Vision-Language Models (VLMs)~\cite{Ghosh2024exploring,Chen2024evolution}, with their holistic feature aggregation, are architecturally incapable of resolving this, frequently misinterpreting the visual patterns of articulation as affective signals, as shown in Figure \ref{fig1}. Second, and equally important, is a critical void in meaningful evaluation. While evaluation protocols are established for affective computing in controlled or semi-controlled settings, they are fundamentally ill-equipped for the fluid, context-dependent nature of in-the-wild dialogue. This leaves the community without a standardized and automated method to assess a generated psychological analysis for its visual grounding, logical depth, and descriptive richness in these challenging, realistic scenarios.

In this work, we tackle these twin challenges of modeling and evaluation head-on. We introduce Multilevel Insight Network for Disentanglement(MIND), a novel hierarchical vision-language architecture designed to explicitly disentangle the visual features of speech from those of emotion. Critically, to measure its true capabilities, we also propose what is, to our knowledge, the first automatic evaluation framework:Psychological Reasoning Insight Scoring Metric(PRISM) specifically designed for the multifaceted assessment of generative psychological analysis in in-the-wild conversational settings. By combining a sophisticated model with a bespoke, meaningful benchmark, we create a complete ecosystem for advancing research in this challenging domain.

To summarize, we make the following contributions:
(1) We propose MIND, a novel hierarchical framework that explicitly distinguishes articulatory movements from emotional signals.
(2) We design and present a comprehensive evaluation framework, PRISM, which is the first of its kind for automatic, vision-based psychological analysis, particularly for natural conversation scenarios.
(3) We construct a new dataset, ConvoInsight-DB, which features macro-expressions, micro-expressions, character emotion analysis, and character psychological analysis.

\definecolor{lightblue}{rgb}{0.93, 0.95, 1.0} 
\definecolor{lightgreen}{rgb}{0.90, 1.0, 0.90} 

\newcolumntype{C}[1]{>{\centering\arraybackslash\hspace{0pt}}p{#1}}

\begin{table*}[!t]
\fontsize{8}{8}\selectfont
\setlength{\tabcolsep}{1.4mm}
\vspace{-3mm}
\rowcolors{2}{lightblue}{white} 
\resizebox{\textwidth}{!}{ 
\begin{tabular}{lcclccc}
  \toprule
  \bf Benchmark&\bf Granularity&\bf Mental activity&\bf Modality&\bf Macro-expression analysis&\bf Micro-expression detection&\bf Micro-expressions analysis\\
  \midrule
  CR~\cite{Blitzer2007biographies} & Coarse & \textcolor{ired}{\ding{55}} & Text & \textcolor{ired}{\ding{55}} & \textcolor{ired}{\ding{55}} & \textcolor{ired}{\ding{55}}\\
  Yelp~\cite{Tang2015document} & Coarse& \textcolor{ired}{\ding{55}} & Text & \textcolor{ired}{\ding{55}} & \textcolor{ired}{\ding{55}} & \textcolor{ired}{\ding{55}}\\
  SemEval~\cite{Pontiki2014semeval} & Fine & \textcolor{ired}{\ding{55}} & Text & \textcolor{igreen}{\ding{51}} & \textcolor{ired}{\ding{55}} & \textcolor{ired}{\ding{55}}\\
  TOWE~\cite{Fan2019target} & Fine & \textcolor{ired}{\ding{55}} & Text &\textcolor{ired}{\ding{55}} & \textcolor{ired}{\ding{55}} & \textcolor{ired}{\ding{55}}\\
  ACOS~\cite{Cai2021aspect} & Fine & \textcolor{ired}{\ding{55}} & Text & \textcolor{igreen}{\ding{51}} & \textcolor{ired}{\ding{55}} & \textcolor{ired}{\ding{55}}\\
  ASTE~\cite{Peng2020knowing} & Fine &\textcolor{ired}{\ding{55}} & Text & \textcolor{igreen}{\ding{51}} & \textcolor{ired}{\ding{55}} & \textcolor{ired}{\ding{55}}\\
  DiaASQ~\cite{Li2023diaasq} & Fine &\textcolor{ired}{\ding{55}} & Text & \textcolor{igreen}{\ding{51}} & \textcolor{ired}{\ding{55}} & \textcolor{ired}{\ding{55}}\\
  VER~\cite{huang2025emotionqwentraininghybridexperts} & Fine & \textcolor{ired}{\ding{55}} & Text, Video & \textcolor{igreen}{\ding{51}} & \textcolor{ired}{\ding{55}} & \textcolor{ired}{\ding{55}}\\
  CMU-MOSEI~\cite{Zadeh2018multimodal} & Coarse & \textcolor{ired}{\ding{55}} & Text, Audio, Video & \textcolor{ired}{\ding{55}} & \textcolor{ired}{\ding{55}} & \textcolor{ired}{\ding{55}}\\
  IEMOCAP~\cite{Busso2008iemocap} & Coarse & \textcolor{ired}{\ding{55}} & Text, Audio, Video & \textcolor{ired}{\ding{55}} & \textcolor{ired}{\ding{55}} & \textcolor{ired}{\ding{55}}\\
  MELD~\cite{Poria2019meld} & Coarse & \textcolor{ired}{\ding{55}} & Text, Audio, Video & \textcolor{ired}{\ding{55}}& \textcolor{ired}{\ding{55}} & \textcolor{ired}{\ding{55}}\\
  M3ED~\cite{Zhao2022m3ed} & Coarse & \textcolor{ired}{\ding{55}} & Text, Audio, Video & \textcolor{ired}{\ding{55}}& \textcolor{ired}{\ding{55}} & \textcolor{ired}{\ding{55}}\\
  CAS(ME)$^2$~\cite{Qu2016casme2} & Coarse & \textcolor{ired}{\ding{55}} & Video & \textcolor{ired}{\ding{55}} & \textcolor{igreen}{\ding{51}} & \textcolor{ired}{\ding{55}}\\
  PanoSent~\cite{luo2024panosent} & Fine & \textcolor{ired}{\ding{55}} & Text, Image, Audio, Video& \textcolor{igreen}{\ding{51}}& \textcolor{ired}{\ding{55}} & \textcolor{ired}{\ding{55}}\\
  \hdashline
  \rowcolor{white}
  \bf{\texttt{PRISM}} & Fine & \textcolor{igreen}{\ding{51}} & Text, Audio, Video & \textcolor{igreen}{\ding{51}} & \textcolor{igreen}{\ding{51}} & \textcolor{igreen}{\ding{51}}\\
 \bottomrule
\end{tabular}
} 
\caption{Summary of existing popular benchmarks of sentiment analysis (representatively summarized, not fully covered).}
\label{tab1}
\vspace{-2mm}
\end{table*}

\section{Related Work}
\label{Related Work}
Our research is positioned at the confluence of VLMs and fine-grained Facial Affective Computing, addressing critical gaps in both representation learning and evaluation.

\textbf{Vision-Language Models and Facial Affective Computing}. The current paradigm in video understanding is dominated by large VLMs like Video-LLaMA~\cite{Zhang2023video} and Qwen2.5-VL~\cite{Qwen2.5-VL}, which couple a holistic visual encoder with a Large Language Model (LLM). Concurrently, the field of Facial Affective Computing has a rich history of developing specialized models for recognizing discrete emotions or detecting micro-expressions ~\cite{Singh2025unveiling,Akilandasowmya2024emotion}. However, both lines of research fall short of our goal. VLMs are architecturally blind to the fleeting, localized facial cues, while traditional affective computing models are framed as recognition tasks, outputting mere labels rather than the rich, inferential generative analysis required to explain the 'why' behind an expression. Our work bridges this divide, reformulating the task from recognition to generative inference, enabled by a fine-grained visual encoder.

\textbf{Disentangled Representation for Facial Analysis}. The principle of learning disentangled representations has proven effective for separating static facial factors like identity from expression~\cite{wang2025fantasyportrait}. This body of work inspires our approach but has yet to address the core challenge of in-the-wild conversations: disentangling two dynamic, co-occurring, and visually ambiguous processes—the visual patterns of articulation and the subtle signals of affect. Therefore, to address this pain point, we designed the MIND visual network to decouple the visual features of pronunciation from the visual features of facial expression changes.

\textbf{Micro-expression information characteristics are linked to human psychological activities}. Previous research has not fully explored the role of micro-expressions in large-scale models' inference of human psychology. In most cases, micro-expression information features are considered as part of macro-expression features. Furthermore, micro-expressions are rarely combined with the specific psychological activities of the characters.Micro-expression information is a key source of information for analyzing a person's true inner thoughts~\cite{Singh2025unveiling2}. To address this, this paper proposes a new dataset, ConvoInsight-DB, designed to bridge these gaps and improve the ability of large-scale models to infer human psychology.

\textbf{Evaluation of Psychological Analysis in Conversational Settings}. A significant bottleneck hindering progress is the lack of meaningful evaluation. While evaluation protocols exist for traditional affective computing—such as F1 scores for emotion classification on datasets like AffectNet~\cite{Mollahosseini2017affectnet}—these are designed for tasks with discrete, unambiguous labels, often in controlled or semi-controlled environments. Standard text generation metrics (e.g., BLEU~\cite{Papineni2002bleu}, ROUGE~\cite{Lin2004rouge}) are inadequate because they only measure superficial n-gram overlap and fail to capture semantic accuracy, factual basis, or psychological depth. They lack the capacity to evaluate the inferential and compositional nature of analysis required for in-the-wild conversations, where an expression's meaning is heavily context-dependent and intertwined with speech. More recent "LLM-as-a-judge" approaches~\cite{Szymanski2025limitations,Wang2025can}, while powerful, cannot verify visual grounding and are not tailored to the specific psychological constructs we aim to measure. Table \ref{tab1} summarizes the key differences between this work and existing benchmarks. Our work directly fills this gap, introducing PRISM, for generative analysis of human psychology in unconstrained conversational contexts.

\begin{figure*}[t]
\centering
\includegraphics[width=0.9\textwidth]{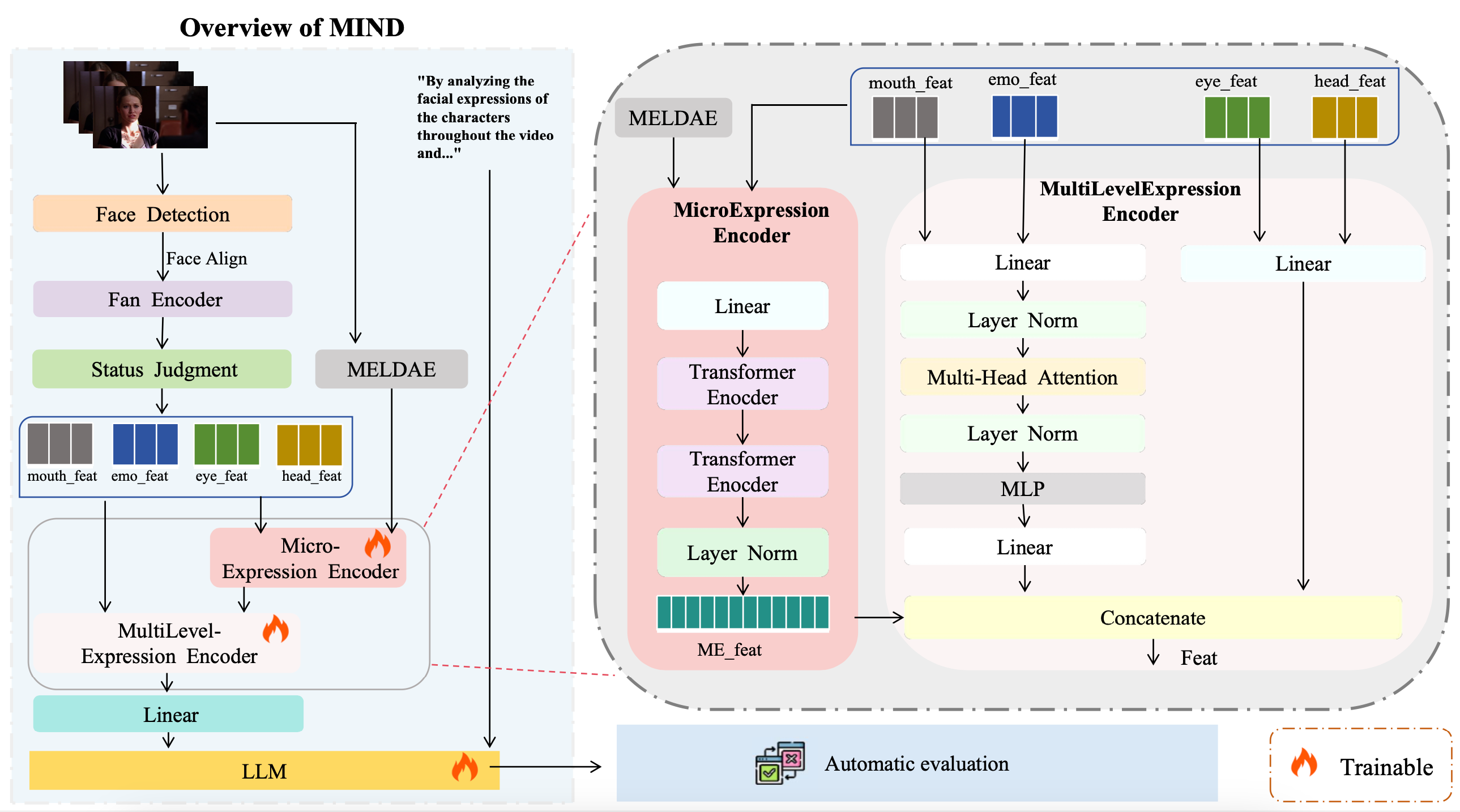} 
\caption{An overview of the MIND training process and model architecture. The FanEncoder module decouples expression features from the video, while the MicroExpressionEncoder extracts micro-expression features. The MultiLevelExpressionEncoder integrates micro-expression emotion features with macro-expression features. The detailed structure of the MicroExpressionEncoder and MultiLevelExpressionEncoder are shown in the figure to the right.}
\label{fig2}
\end{figure*}

\section{Method}
We propose \texttt{MIND}, whose overall architecture is shown in Figure \ref{fig2}. It consists of three main stages: (1) Spatiotemporal Feature Decomposition and Disentanglement; (2) The \texttt{MIND} Core Encoder for parallel, multi-level encoding; and (3) Vision-Language Alignment and Generative Reasoning.

\subsection{Spatiotemporal Feature Decomposition and Disentanglement}
\label{sec:Spatiotemporal Feature Decomposition and Disentanglement}

This stage, corresponding to the left flow in Figure \ref{fig2}, aims to: (1) extract a set of structured, multi-dimensional facial dynamics features from the raw video; and (2) perform our key disentanglement step to resolve Articulatory-Affective Ambiguity.

\paragraph{Expert Feature Extraction}
Given a raw video clip $V$, we first utilize a set of pre-trained, frozen expert models to decompose it into parallel, semantically-disentangled feature streams. This flow includes three sub-steps:
\begin{itemize}
    \item \textbf{Face Detection~\cite{huang2020curricularface}:} Localizes facial bounding boxes in each frame.
    \item \textbf{Face Align~\cite{wang2022pdfgc}:} Normalizes the detected facial regions to mitigate scale and rotation effects.
    \item \textbf{Dynamics Encoding:} Feeds the aligned facial sequence into the \texttt{Fan Encoder}~\cite{Wang2023progressive}, which explicitly decomposes the complex facial dynamics into four key feature streams:
    \begin{itemize}
        \item \textbf{Head Pose ($E_{head}$):} \texttt{head\_feat}
        \item \textbf{Eye Dynamics ($E_{eye}$):} \texttt{eye\_feat}
        \item \textbf{Core Emotion ($E_{emo}$):} \texttt{emo\_feat}
        \item \textbf{Lip Movement ($E_{lip}$):} \texttt{mouth\_feat}
    \end{itemize}
\end{itemize}

\paragraph{Context-Aware Disentanglement}
This is the key innovation in our methodology for resolving Articulatory-Affective Ambiguity. After feature extraction, we introduce the \texttt{Status Judgment} module, which operates exclusively on the most ambiguous $E_{lip}$ stream.

Principle: Our core hypothesis is that the articulation state visually manifests as continuous and high-amplitude variations in the \texttt{mouth\_feat} feature over time, whereas silent or simple emotional expressions (e.g., holding a smile) produce features that are comparatively static or change only briefly.

Implementation: The \texttt{Status Judgment} module performs a Temporal Contrastive Test on the $E_{lip}$ feature sequence $\mathbf{f}_{1...T}$ (where $\mathbf{f}_t \in \mathbb{R}^{D_{lip}}$). Specifically, we assess the magnitude and continuity of its variation by computing the temporal variance ($V_{var}$) and the sum of adjacent differences ($V_{sad}$), as defined in Equation (1):

\begin{equation}
c = \mathbb{I} \left( \alpha \cdot V_{var} + \beta \cdot V_{sad} > \tau \right)
\label{eq:status_judgment}
\end{equation}
where $V_{var}$ and $V_{sad}$ are defined as:
\begin{equation}
V_{var} = \frac{1}{T} \sum_{t=1}^{T} ||\mathbf{f}_t - \boldsymbol{\mu}||_2^2 \quad \text{and} \quad V_{sad} = \sum_{t=2}^{T} ||\mathbf{f}_t - \mathbf{f}_{t-1}||_2
\label{eq:var_sad}
\end{equation}
Here, $\boldsymbol{\mu}$ is the mean feature vector over all frames, and $\mathbb{I}(\cdot)$ is the indicator function. $c=1$ indicates an "articulation state," and $c=0$ indicates a "non-articulation state." $\alpha, \beta,$ and $\tau$ are hyperparameters.

Purification: Based on this judgment, we perform a gated suppression on the \texttt{mouth\_feat} stream to obtain the purified lip feature $\hat{E}_{lip}$:
\begin{equation}
\hat{E}_{lip} = (1 - c) \cdot E_{lip}
\label{eq:purification}
\end{equation}
This step ensures that the subsequent \texttt{MIND} core encoder is not misled by purely articulatory motions, achieving noise disentanglement at the source of the visual modality.

\begin{figure}[t]
\centering
\includegraphics[width=1\linewidth]{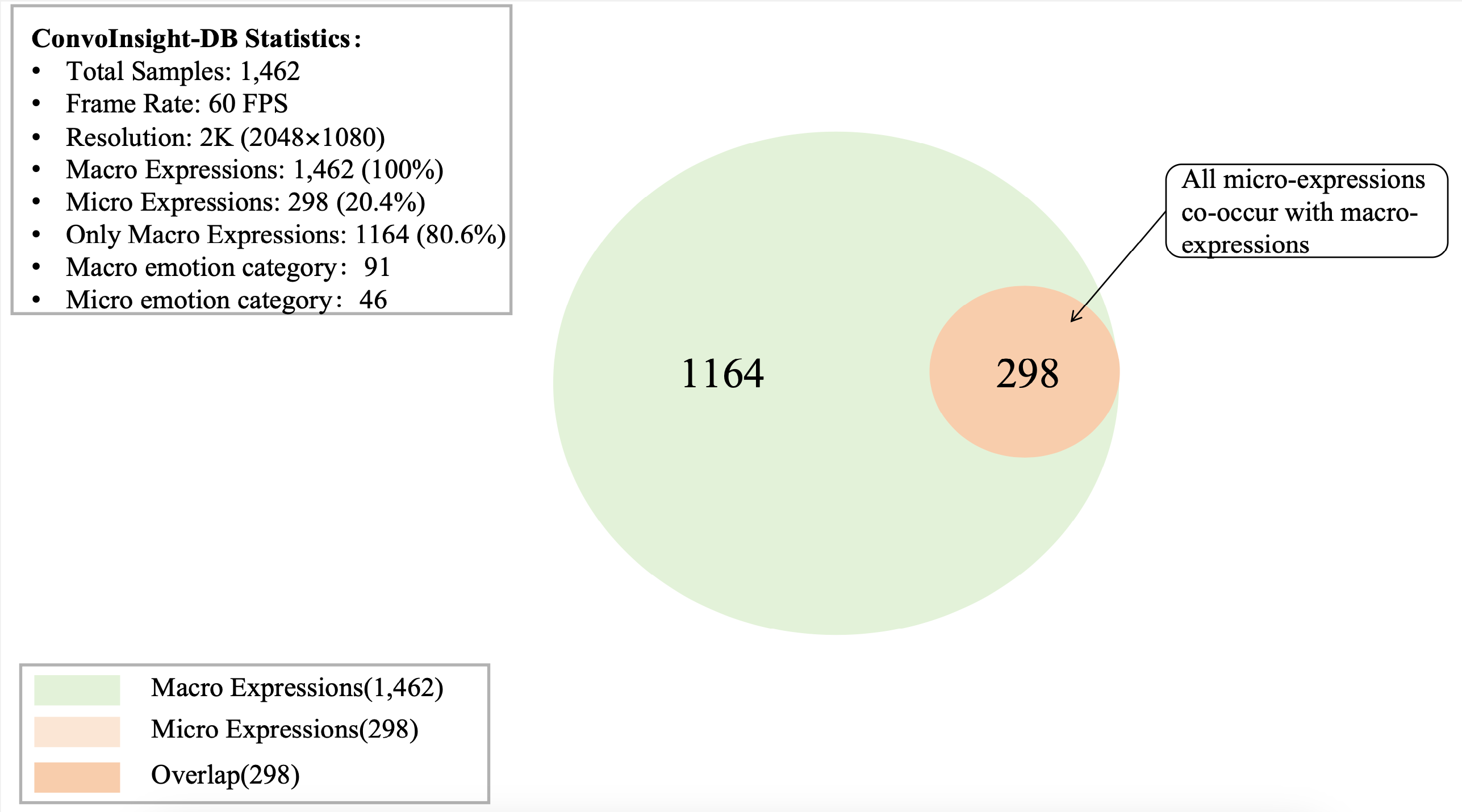}
\caption{Main statistics of ConvoInsight-DB dataset.}
\label{fig3}
\vspace{-5pt}
\end{figure}

\subsection{The MIND Core Encoder}
\label{sec:The MIND Core Encoder}
This is the core innovation of our method, corresponding to the right-hand detail in Figure \ref{fig2}. The \texttt{MIND} framework receives the (purified) feature streams from Step 1 and processes them in a parallel, multi-level manner to produce a single, information-rich visual representation. It consists of two parallel sub-modules: the \texttt{MicroExpressionEncoder} and the \texttt{MultiLevelExpressionEncoder}.

\paragraph{Transient Path: MicroExpressionEncoder}

This module (the pink box in Figure \ref{fig2}) is specialized in capturing fleeting yet critical micro-expression signals, guided by explicit temporal localization.

\begin{itemize}
    \item \textbf{Inputs:} The module receives two distinct sets of inputs:
        \begin{enumerate}
            \item \textbf{Feature Streams:} $E_{\mathrm{head}}, E_{\mathrm{eye}}, E_{\mathrm{emo}}, \hat{E}_{\mathrm{lip}}$.
            \item \textbf{Temporal Localization:} Start and end timestamps ($t_{\mathrm{start}}, t_{\mathrm{end}}$) identified by a parallel \texttt{MELDAE}~\cite{feng2025} module.
        \end{enumerate}
    \item \textbf{Processing:} The module extracts the segment $S_{ME} = [E_{\mathrm{head}}, E_{\mathrm{eye}}, E_{\mathrm{emo}}, \hat{E}_{\mathrm{lip}}]_{t_{\mathrm{start}}:t_{\mathrm{end}}}$.
    \item \textbf{Output:} Generates a compact feature $F_{ME} \in \mathbb{R}^{D_{me}}$.
\end{itemize}

\paragraph{Fusion Path: MultiLevelExpressionEncoder}
This module (the light pink box in Fig. 2) is the "fusion center" of the entire system, and its task is to enhance facial features. It processes features from different levels of abstraction in parallel and intelligently combines them.
\begin{itemize}
    \item \textbf{Input:} The module receives five parallel inputs: $F_{ME}$, $\hat{E}_{lip}$,$E_{emo}$, $E_{eye}$,$E_{head}$ 
    \item \textbf{Processing:} The module features four parallel processing paths corresponding to different feature hierarchies:
    \begin{itemize}
        \item \textbf{Path 1 (Transient):} $F_{ME}$ is treated as the highest-level feature and is passed directly to the final fusion step without further processing.
        \item \textbf{Path 2 (Complex Dynamics):} $\hat{E}_{lip}$ and $E_{emo}$, the two most complex expression features, are first concatenated. They are then fed into a deep processing stream: \texttt{Linear} $\rightarrow$ \texttt{Layer Norm} $\rightarrow$ \texttt{Multi-Head Attention} $\rightarrow$ \texttt{Layer Norm} $\rightarrow$ \texttt{MLP} $\rightarrow$ \texttt{Linear}. This path allows the model to fully mine the complex relationships between the lip and core emotion features.
        \item \textbf{Path 3 (Simple Dynamics - Eye):} $E_{eye}$ is passed through a separate \texttt{Linear} layer for simple feature mapping.
        \item \textbf{Path 4 (Simple Dynamics - Head):} $E_{head}$ is similarly passed through a separate \texttt{Linear} layer.
    \end{itemize}
    \item \textbf{Fusion:} As depicted, the output features from all four paths ($F'_{ME}, F'_{complex}, F'_{eye}, F'_{head}$) are finally concatenated at the \texttt{Concatenate} step to form a single, comprehensive fusion vector $V_{MIND}$ (\texttt{Feat} in the figure):
\end{itemize}
\begin{equation}
V_{MIND} = \text{Concat}(F'_{ME}, F'_{complex}, F'_{eye}, F'_{head})
\label{eq:fusion}
\end{equation}

\subsection{Vision-Language Alignment and Generative Reasoning}
\label{sec:Vision-Language Alignment and Generative Reasoning}
This final stage corresponds to the last part of the flowchart, responsible for translating our distilled visual evidence $V_{MIND}$ into natural language analysis that the LLM can understand and reason with.
\begin{itemize}
    \item \textbf{Visual Alignment:} The fused feature vector $V_{MIND}$ is first mapped into the LLM's word embedding space via a \texttt{Linear} layer (the projector $W_p$) to obtain $V_{proj} \in \mathbb{R}^{D_{llm}}$.
    \item \textbf{Instruction Injection \& Generation:} $V_{proj}$ is injected into the \texttt{LLM} to replace a special visual token \texttt{<expr>} in the prompt template.
\end{itemize}
This step can be formulated as follows: 

\vspace{-2mm}
\begin{tcolorbox}[fontupper=\small]
\vspace{-2mm}
{\small
\textbf{Input Data}: \(V_{MIND}\) \\
\textbf{Instruction}:You are an expert in psychology and micro-expression analysis. Please provide a detailed analysis of the person's facial expressions, emotional state, and inner psychology based on this expression feature.
\\

\textbf{\color{blue}{Expected Output}}: Analysis Text\\
}
\vspace{-4mm}
\end{tcolorbox}
\vspace{-2mm}

This step can be expressed as:
\begin{equation}
Analysis_{Text} \leftarrow f_{LLM}(P_{prompt} \oplus W_p(V_{MIND}))
\label{eq:generation}
\end{equation}
where $f_{LLM}$ is the Large Language Model, which we fine-tune using Parameter-Efficient Fine-Tuning (PEFT) with Low-Rank Adaptation (LoRA). The $\oplus$ operation denotes the injection of the projected visual vector $V_{proj}$ to replace the \texttt{<expr>} token's embedding. Finally, the generated $Analysis_{Text}$ is fed into our \texttt{Automatic evaluation} framework for assessment.

\subsection{Automatic Evaluation Framework}
\label{sec:evaluation}
PRISM leverages the powerful external large-scale language model GPT-4o~\cite{Hurst2024gpt4o}, enabling it to act as a "psychology expert." For each video sample, LLM assessors receive detailed scoring guidelines across four dimensions set by psychologists: ground-truth "micro-expression labels" (out of 1.25 points), ground-truth "macro-expression labels" (out of 1.5 points), psychological insight and reasoning depth (out of 1.25 points), and detail coverage and richness (out of 1 point). The final score for each generative analysis is calculated as the sum of these four dimensions (out of 5). This provides a robust, reproducible, and semantically rich benchmark for our task. The complete and detailed scoring criteria provided to LLM evaluators can be found in the Appendix C.

\begin{figure}[t]
\centering
\includegraphics[width=1\linewidth]{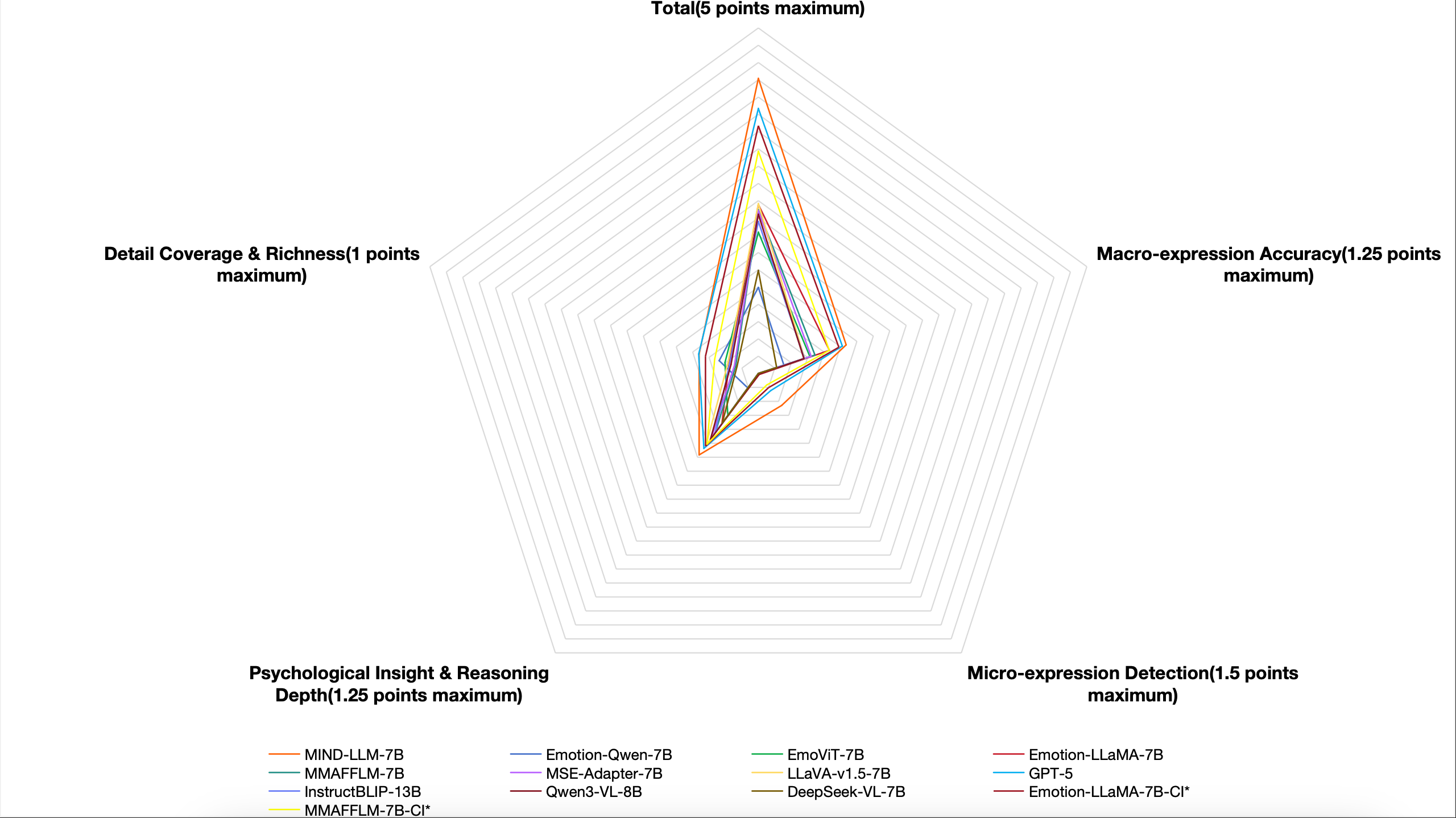}
\caption{Balanced performance of large multimodal models (LMMs).MIND-LLM-8B demonstrates superior performance across all evaluation dimensions. Evaluation details are shown in Table \ref{tab3}.}
\label{fig4}
\vspace{-5pt}
\end{figure}

\section{Experiments}
\subsection{ConvoInsight-DB}
The ConvoInsight-DB dataset, used to analyze sentiment in natural dialogue scenarios, was primarily collected by crawling a large number of publicly available movies and TV shows.The focus on micro-expression clips from open-source movies is intended to lay the groundwork for our future micro-expression video generation model; the data we are currently using will be used to guide our micro-expression video generation model. We then conducted a rigorous screening process (combining manual review with automated filtering, such as keyword detection and Toxic-BERT detection~\cite{Detoxify}) to eliminate instances that were harmful, personal, or emotionally irrelevant. After obtaining the cleaned corpus, we collaborated with a team of psychology experts to label the video data. To ensure reliability, each video was independently annotated by at least three different psychology professionals, as well as an AI psychology expert (role-playing). After the annotations were completed, three different psychology professors discussed and jointly finalized the labels. Instances with unresolved ambiguities were removed. Figure \ref{fig3} shows the details of ConvoInsight-DB. 

\subsection{Implementation Details}
Following the pre-training process, MIND uses eight NVIDIA H800 80GB GPUs for distributed training. The training strategy employs the LoRA efficient fine-tuning method, with rank set to 16, alpha set to 32, and dropout set to 0.05. See Appendix B for specific implementation details.

\subsection{ConvoInsight-DB dataset validity verification}
To assess the effectiveness of the dataset, we conducted a comprehensive evaluation. We fine-tuned the Qwen2.5vl-7b~\cite{Qwen2.5-VL} model (CIQwen-7b) using the ConvoInsight-DB dataset and compared CIQwen-7b with the GPT4o~\cite{Hurst2024gpt4o} direct role-playing baseline method using the PRISM. As shown in Table.\ref{tab2}, CIQwen-7b consistently and significantly outperformed the baseline method in all dimensions: Macro-expression Accuracy (+23.5\%), Micro-expression Detection (+83.3\%), Psychological Insight \& Reasoning Depth( +31.6\%), and Detail Coverage \& Richness (+48\%). This clearly demonstrates the effectiveness of our dataset in enhancing the model's sentiment analysis capabilities.

\begin{table}[ht]
\centering
\small
\begin{tabular}{llccc}
\toprule
Method  & Mac & Mic & PIRD & DCR \\
\midrule
Role-play & 0.51 & 0.06 &0.57& 0.25 \\
CIQwen-7b & 0.63 &0.11 & 0.75 & 0.37 \\
\midrule
\textbf{Improvement}  & \textbf{+23.5\%} & \textbf{+83.3\%} & \textbf{+31.6\%} & \textbf{+48\%} \\
\bottomrule
\end{tabular}
\caption{Comparison of Role-play vs. CIQwen-7b across emotion categories. Metric abbreviations: Mac (Macro-expression Accuracy), Mic (Micro-expression Detection), PIRD (Psychological Insight \& Reasoning Depth), DCR (Detail Coverage \& Richness).}
\label{tab2}
\vspace{0.1cm}
\end{table}

\subsection{Comparison with Baselines}
\textbf{Baselines and Metrics}. We selected several large-scale models focused on video sentiment analysis and general large-scale models as baselines for comparative evaluation, including Emotion-Qwen-7B~\cite{huang2025emotionqwentraininghybridexperts}, EmoViT-7B~\cite{Xie2024EmoVIT}, Emotion-LLaMA-7B~\cite{NEURIPS2024_c7f43ada}, MMAFFLM-7B~\cite{liu2025mmaffben}, MSE-Adapter-7B~\cite{Yang2025mse}, LLaVA-v1.5-7B~\cite{liu2023llava},GPT-5~\footnote{\url{https://chatgpt.com/}}, InstructBLIP-13B~\cite{li2022lavis}, Qwen3-VL-8B~\cite{Qwen2025qwen3vl}, and DeepSeek-VL-7B~\cite{lu2024deepseekvl}. The LLM in our MIND uses the Qwen3-8B~\cite{Yang2025qwen3} model(MIND-LLM-8B). We evaluate all methods on PRISM.See Appendix A for model prompts.

\textbf{Quantitative Results}. Table \ref{tab3} compares the performance of MIND-LLM-8B with other baseline models on the PRISM evaluation framework, yielding the following observations. First, due to the rich micro-expression features in the data, we found that both general-purpose large models and sentiment analysis large models performed very poorly, failing to recognize almost any micro-expressions. However, enhancing sentiment understanding and inference capabilities can address this issue to some extent, thereby improving overall performance. For example, compared to the best-performing baseline model (GPT-5), our MIND-LLM-8B achieved an 86.95\% performance improvement in micro-expression detection. Interestingly, we found that GPT-5 performed best among all baseline models, which was unexpected, indicating that larger parameters do indeed improve the model's understanding capabilities. For fairness, we also fine-tuned the two best-performing sentiment analysis large models, Emotion-LLaMA-7B and MMAFFLM-7B, using our ConvoInsight-DB dataset (Emotion-LLaMA-7B-CI*, MMAFFLM-7B-CI*). It is evident that Emotion-LLaMA-7B-CI and MMAFFLM-7B-CI received good feedback in each dimension, almost comparable to GPT-5, but still slightly inferior to MIND-LLM-8B. Overall, MIND-LLM-8B achieved significant results in facial expression analysis and psychological interpretation.

\begin{table}[ht]
\centering
\small
\begin{tabular}{lcccc}
\toprule
Model & \multicolumn{4}{c}{Metrics} \\
\cmidrule{2-5}
 & Mac & Mic & PIRD & DCR\\
\midrule
\textbf{Emotion LMMs}            & & & & \\
Emotion-Qwen-7B          & 0.233 & 0 & 0.156 & 0.358 \\
EmoViT-7B                & 0.471 &0 & 0.415 & 0.305 \\
Emotion-LLaMA-7B         & 0.649 & 0.013 & 0.538 & 0.268 \\
MMAFFLM-7B               & 0.517 & 0.006 & 0.636 & 0.207 \\
MSE-Adapter-7B           & 0.482 & 0.002 &0.692  & 0.219 \\
Emotion-LLaMA-7B-CI*      & 0.735 & 0.152 & 0.781 & 0.483 \\
MMAFFLM-7B-CI*            & 0.648 & 0.124 & 0.759 & 0.399 \\
\midrule 
\textbf{General LMMs}             & & & & \\
GPT-5                    & 0.767 & 0.184 & 0.805 & 0.545\\
LLaVA-v1.5-7B            & 0.422 & 0 & 0.774 &0.279\\      
InstructBLIP-13B         & 0.420 & 0 & 0.637  & 0.263 \\
Qwen3-VL-8B              & 0.416 & 0.009 & 0.708 & 0.250 \\   
DeepSeek-VL-7B           & 0.167 &0 & 0.533 & 0.195 \\                      
\midrule
\textbf{MIND-LLM-8B}             & \textbf{0.802} &\textbf{0.344} & \textbf{0.875} & \textbf{0.541} \\
\bottomrule
\end{tabular}
\caption{Comparison of MIND-LLM-8B with other large emotion models and general large models on PRISM.Metric abbreviations: Mac (Macro-expression Accuracy), Mic (Micro-expression Detection), PIRD (Psychological Insight \& Reasoning Depth), DCR (Detail Coverage \& Richness).}
\label{tab3}
\end{table}

Finally, we can examine the task evaluation results from different perspectives. For micro-expression detection, current mainstream emotion models all perform well. However, it can be seen that nearly all models perform significantly worse in this area. Recognizing and reasoning micro-expressions is the greatest challenge facing AI in understanding human psychology, providing a challenging benchmark for subsequent research.

\subsection{Ablation Studies}%
\label{sec:ablation}

To rigorously validate the effectiveness of our proposed \texttt{MIND} framework and quantify the contribution of each of its core components, we conduct a comprehensive series of simplification studies. We systematically degrade the performance of the full model by removing or replacing key modules and evaluate these variants using PRISM, our proposed automated evaluation framework. The results are summarized in Table \ref{tab4}.

\begin{table}[t]
\centering
\small
\begin{tabular}{@{}lcccc@{}}
\toprule
\textbf{Variant} & \textbf{Mac} & \textbf{Mic} & \textbf{PIRD} & \textbf{DCR} \\
\midrule
\textbf{MIND (Full)} & \textbf{0.802} & \textbf{0.344} & \textbf{0.875} & \textbf{0.541} \\
\hline
(a) w/o Status Judgment & 0.739 & 0.286 & 0.727 & 0.425  \\
(b) w/o Micro.Enc & 0.718 & 0.302 & 0.760 & 0.441  \\
(c) w/o Multi.Enc & 0.653 & 0.228 & 0.713 & 0.391  \\
(d) Basic Fusion (MLP)& 0.730 & 0.236 & 0.656 & 0.309  \\
\bottomrule
\end{tabular}
\caption{Ablation studies were conducted using the ConvoInsight-DB dataset on PRISM. Metric abbreviations: Mac (Macro-expression Accuracy), Mic (Micro-expression Detection), PIRD (Psychological Insight \& Reasoning Depth), DCR (Detail Coverage \& Richness).}
\label{tab4}
\vspace{0.1cm}
\end{table}

\begin{table}[h]
\centering
\begin{tabular}{@{}lcc@{}}
\toprule
\textbf{LLM Backbone} & \textbf{Size} & \textbf{Final Score} \\
\midrule
ViT + Qwen3-8B & 8B & 1.496  \\
\midrule
MIND + Llama-3-8B & 8B & 2.419 \\
\textbf{MIND + Qwen3-8B} & \textbf{8B} & \textbf{2.562} \\
\bottomrule
\end{tabular}
\caption{Comparison of different LLM backbones integrated with our frozen MIND visual encoder.}
\label{tab5}
\end{table}

\paragraph{Effectiveness of the  Status Judgment Module.}
To verify our core disentanglement hypothesis, we create a variant (a) \textbf{w/o  Status Judgment}. In this setting, we completely bypass the module described in Sec 3.1.2. The raw, unpurified \texttt{mouth\_feat} ($E_{lip}$) is directly fed into the subsequent encoders. As shown in Table \ref{tab4}, the performance of this variant is severely degraded, with Micro-expression Detection performance decreasing by 16.86\%. The impact is most devastating on the "Micro-expression Detection" and "Psychological Insight" dimensions. This empirically validates our central claim: without explicitly identifying and suppressing articulatory noise, the model is fatally misled by speech-related motions. This leads to factually incorrect visual grounding (mistaking speech for emotion) and, consequently, logically flawed psychological inferences.

\paragraph{Effectiveness of the MicroExpressionEncoder Module.}
To quantify the contribution of our specialized transient path, we design variant (b) \textbf{w/o Micro.Enc} Here, we remove the entire \texttt{MicroExpressionEncoder} and its corresponding path. The \texttt{MultiLevelExpressionEncoder} thus loses its most abstract input, $F_{ME}$, and must infer the entire state solely from the other four feature streams. Table \ref{tab4} shows a significant performance drop, primarily concentrated in the "Micro-expression Detection" score, which plummets from 0.344 to 0.302. This confirms that a dedicated, attentive module is crucial for isolating the faint, transient signals of micro-expressions, which are otherwise lost or "averaged out" by the main fusion encoder.

\paragraph{Importance of the Micro-Expression Encoder.}
To verify the expression enhancement effect of the MultiLevelExpressionEncoder block, we created a variant (c) \textbf{w/o MultiL.Enc}. This variant directly inputs features into the subsequent Liner layer without the feature enhancement provided by the MultiLevelExpressionEncoder module. As shown in Table \ref{tab4}, this variant exhibits a significant performance decrease, with a 18.58\% drop in Macro-expression Accuracy and a 31.40\% drop in Micro-expression Detection. This effectively demonstrates the necessity of the MultiLevelExpressionEncoder module.

\paragraph{Contribution of the MultiLevel-Expression Encoder.}
To validate our hierarchical fusion architecture, we test variant (d) \textbf{Basic Fusion (MLP)}. We compares its \texttt{Hybrid Compressor} against a simpler \texttt{MLP projector}. Here, we replace our complex, parallel-path \texttt{MultiLevelExpressionEncoder} (the light pink box in Figure \ref{fig2}) with a simple "brute-force" fusion block: all five input features ($F_{ME}$, $\hat{E}_{lip}$, $E_{emo}$, $E_{eye}$, $E_{head}$) are first concatenated and then passed through a single, deep MLP to produce the final $V_{MIND}$. Table \ref{tab4} show that this non-hierarchical approach underperforms our full model across all four dimensions. This validates our design of using specialized, parallel paths (e.g., the deep attention stream for $E_{emo}$/$\hat{E}_{lip}$ vs. simple \texttt{Linear} layers for $E_{eye}$/$E_{head}$) to handle features of different complexities, proving it is more effective than a monolithic fusion block.

\paragraph{Impact of the LLM Backbone.}
Finally, we explore the generalization ability of our framework by replacing the LLM backbone model. We keep the \texttt{MIND} visual encoder unchanged and retrain the projector and LoRA adapter for an alternative model \texttt{Llama-3-8B}~\cite{Dubey2024llama3} and compare it with the default \texttt{Qwen3-8B}~\cite{Yang2025qwen3}. The results are listed in Table \ref{tab5}. As shown, the more powerful LLM backbone model consistently achieves higher scores, especially on the "psychological insight" dimension, confirming that high-level reasoning capabilities are crucial for the final analysis quality~\cite{Susnjak2024chatgpt}. However, we also observe that our \texttt{MIND} framework significantly outperforms the baseline model across all backbone models, indicating that our visual front-end is a robust, model-agnostic component that can provide critical, cleansed signals to any downstream LLM.

\section{Discussion}
\label{sec:discussion}

Our experimental results, particularly the ablation studies, provide strong evidence for our central hypotheses. However, these results also illuminate several non-trivial challenges, limitations, and avenues for future inquiry.

\paragraph{On the Criticality of Disentanglement.}
The most significant finding from our ablations (Table \ref{tab4}) is the catastrophic performance collapse observed in the "w/o Status Judgment" variant. The final score dropped, which empirically validated our core argument: clearly addressing euphony is not merely a "nice-to-have" feature, but a fundamental prerequisite for accomplishing this task. It suggests that without a mechanism to filter speech-related visual noise at the source, the model is fatally misled, leading to factually incorrect visual grounding. This finding aligns with the broader trend in multimodal research, such as the \texttt{Emotion-Qwen} paper's use of specialized expert modules~\cite{huang2025emotionqwentraininghybridexperts}, which argues against one-size-fits-all processing and in favor of specialized components for distinct sub-tasks.

\paragraph{Reflections on our Evaluation Framework.}
A key contribution of this work is the proposal of a new, fine-grained automatic evaluation framework. Our experiments revealed its utility: baseline models, while sometimes generating plausible-sounding psychological "fluff," scored near-zero on the "Micro-expression Detection" dimension. A standard metric like BLEU or even a generic "LLM-as-a-judge" would have been susceptible to these "plausible hallucinations." Our framework, by explicitly grounding the score in the detection of specific macro/micro expressions, provided a much-needed, verifiable assessment of the model's factual accuracy and reasoning depth. This demonstrates the necessity of domain-specific evaluation protocols for nuanced tasks.

\paragraph{Limitations of the Status Judgment Heuristic.}
While our \texttt{Status Judgment} module (Sec 3.1.2) proved highly effective, it remains a heuristic-based approach. The formula, $c = \mathbb{I}(\alpha \cdot V_{var} + \beta \cdot V_{sad} > \tau)$, is based on the assumption that speech always produces high-variance features. This assumption may falter in edge cases:
\begin{itemize}
    \item \textbf{Subtle Talkers:} A person speaking very quietly or with minimal lip movement might be misclassified as "non-speaking" ($c=0$), causing their articulatory movements to be incorrectly fed into the model as emotional cues.
    \item \textbf{Expressive Non-Speakers:} Conversely, a person exhibiting rapid, non-speech-related facial tics or chewing might be misclassified as "speaking" ($c=1$), causing their genuine (but noisy) expression features to be erroneously suppressed.
\end{itemize}
A more robust, end-to-end learned disentanglement module is a promising direction for future work.

\paragraph{Limitations of the Visual-Only Modality.}
A more fundamental limitation is our model's deliberate reliance on a single modality (vision). By design, \texttt{MIND} is deaf. It cannot distinguish between a subject holding their breath in terror (a purely visual cue) and a subject merely pausing mid-sentence. While our work pushes the boundary of what can be inferred from vision alone, the integration of audio prosody (pitch, tone, energy) is the clear next step to resolve such deep ambiguities and achieve a truly holistic understanding.

\paragraph{Broader Impact and Ethical Considerations.}
As with any technology capable of inferring human psychological states, the potential for misuse must be soberly addressed. An automated system for psychological analysis, especially in non-clinical, "in-the-wild" settings, carries significant ethical risks, including surveillance, privacy violation, and use in biased contexts (e.g., hiring, interrogations). Furthermore, our model's interpretations are inherently tied to the cultural and demographic biases present in its training data; expressions of "sadness" or "confidence" are not universal. We position \texttt{MIND} not as a "lie detector" or a diagnostic tool, but as a research step towards building more empathetic AI assistants (e.g., for mental health chatbots or assistive technology). We strongly advocate for future work in this area to include rigorous auditing for fairness, bias, and transparency before any real-world deployment.

\section{Conclusion}
\label{sec:conclusion}
This paper tackled the critical challenge of Articulatory-Affective Ambiguity in generative psychological analysis of in-the-wild videos. We introduced MIND, a hierarchical visual encoder that explicitly disentangles speech-related motion from emotion, and proposed a novel, multi-dimensional automatic evaluation framework to assess visually-grounded psychological insight. Our experiments demonstrate that MIND significantly outperforms baselines, validating the necessity of explicit visual disentanglement. Together, our model, dataset (ConvoInsight-DB), and evaluation protocol provide a robust foundation for future research in this domain. Code and data will be released to foster reproducibility and further progress. Future work will explore integrating audio modalities for a more holistic understanding.

\bibliographystyle{ieeenat_fullname}
\bibliography{main}


\end{document}